\documentclass[sigconf, 10pt]{acmart}
\usepackage{subcaption}
\usepackage{enumitem}

\AtBeginDocument{%
  }

\begin{document}
\captionsetup[figure]{font=small}
\captionsetup[table]{font=small}

\title{TDiff: Thermal Plug-And-Play Prior with Patch-Based Diffusion}

\newcommand{\TDiff}{{\sc TDiff\ }}
\newcommand{\TDiffVS}{{\sc TDiff-32\ }}
\newcommand{\TDiffS}{{\sc TDiff-64\ }}
\newcommand{\TDiffL}{{\sc TDiff-128\ }}

\newcommand{\TDiffST}{{\sc \textbf{TDiff-64}\ }}
\newcommand{\TDiffLT}{{\sc \textbf{TDiff-128}\ }}

\begin{abstract}
Thermal images from low-cost cameras often suffer from low resolution, fixed pattern noise, and other localized degradations. Available datasets for thermal imaging are also limited in both size and diversity. To address these challenges, we propose a patch-based diffusion framework (\TDiff) that leverages the local nature of these distortions by training on small thermal patches. In this approach, full-resolution images are restored by denoising overlapping patches and blending them using smooth spatial windowing. To our knowledge, this is the first patch-based diffusion framework that models a learned prior for thermal image restoration across multiple tasks. Experiments on denoising, super-resolution, and deblurring demonstrate strong results on both simulated and real thermal data, establishing our method as a unified restoration pipeline.
\end{abstract}



\author{Piyush Dashpute}
\affiliation{%
  \institution{University of California, Riverside}
  \country{United States}
  }
\email{pdash003@ucr.edu} 

\author{Niki Nezakati}
\affiliation{%
  \institution{University of California, Riverside}
  \country{United States}
  }
\email{nneza001@ucr.edu}

\author{Wolfgang Heidrich}
\affiliation{%
  \institution{King Abdullah University of Science and Technology}
  \country{Saudi Arabia}
  }
\email{wolfgang.heidrich@kaust.edu.sa}

\author{Vishwanath Saragadam}
\affiliation{%
  \institution{University of California, Riverside}
  \country{United States}
  }
\email{vishwanath.saragadam@ucr.edu}

\acmYear{2025}\copyrightyear{2025}
\setcopyright{cc}
\setcctype[4.0]{by}
\acmConference[MobiCom '25]{ACM International Workshop on Thermal Sensing and Computing}{November 4--8, 2025}{Hong Kong, China}
\acmBooktitle{ACM International Workshop on Thermal Sensing and Computing (MobiCom '25), November 4--8, 2025, Hong Kong, China}
\acmDOI{10.1145/3737905.3769286}
\acmISBN{979-8-4007-1982-0/25/11}

\maketitle

\section{Introduction}
Thermal imaging captures longwave infrared emission and reflection, enabling applications in surveillance, medical diagnostics, industrial inspection, and material classification~\cite{wilson2023recent, dashpute2023thermal}. Thermal cameras work in darkness, smoke, and fog. Consumer-grade units (uncooled microbolometers) have low resolution (often below 1 MP), limited frame rates, and fixed pattern noise (FPN). Unlike Gaussian noise, FPN is structured and repeatable (e.g., streaks), which standard denoising methods may mistake for real features. These FPN statistics cause deep models trained on visible or near-infrared data to underperform on thermal images~\cite{venkatakrishnan2013plug,romano2017little,zhang2017beyond,zhang2018ffdnet}. Classical methods~\cite{rudin1992nonlinear,donoho2002noising} are training-free but struggle on complex thermal scenes, and recent self-supervised approaches~\cite{ulyanov2018deep,saragadam2021thermal} also fall short. A key challenge is data scarcity, as thermal datasets are far smaller than visible-light counterparts and often comprise correlated video frames, limiting independence~\cite{wilson2023recent}.

\begin{figure}[ht]
  \centering
  \includegraphics[width=\linewidth]{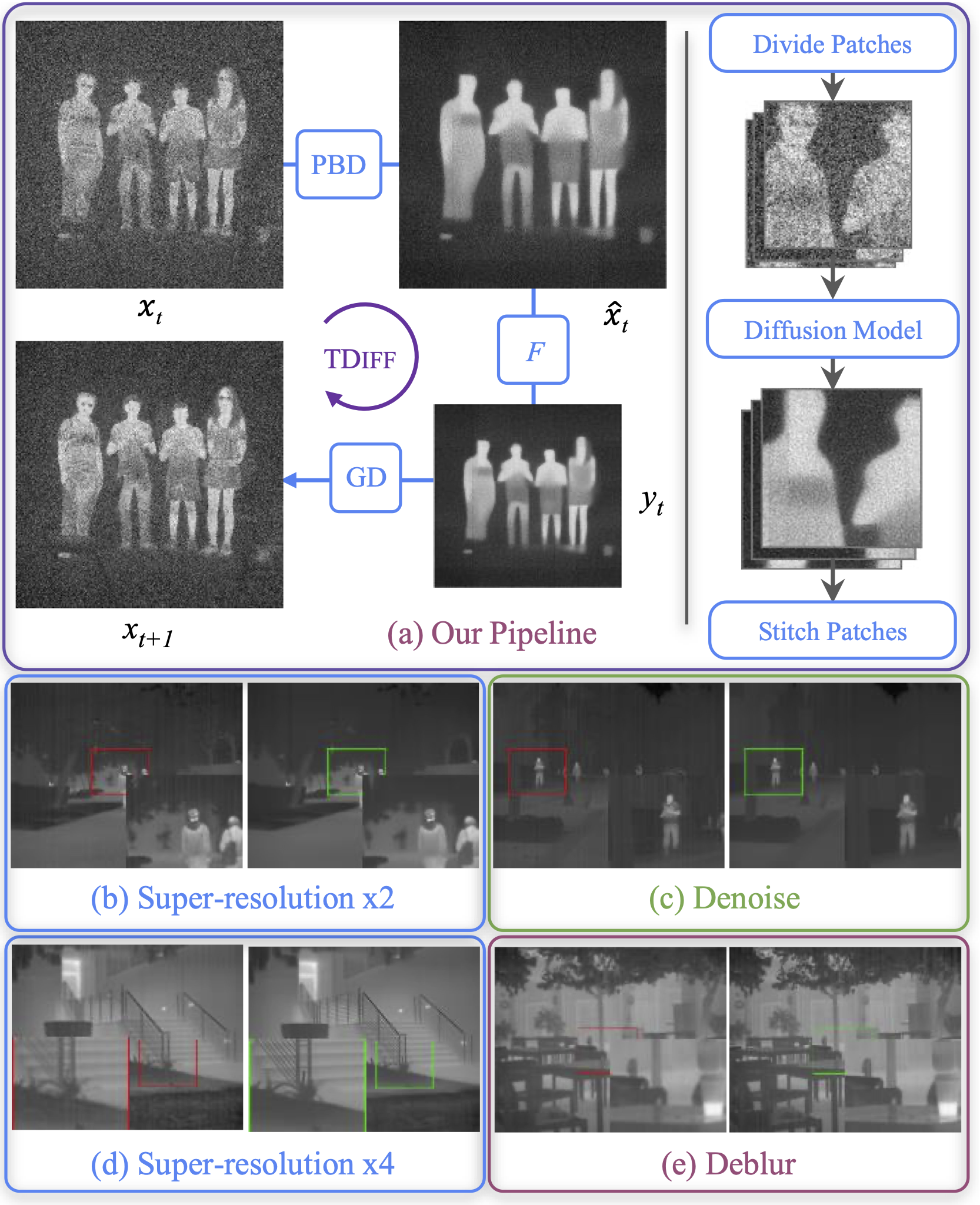}
\small\caption{\textbf{Overview of \TDiff pipeline.} (\textit{Top}) A degraded thermal image $x_t$ is divided into overlapping patches, denoised by our patch-based diffusion model, and recombined into $\hat{x}_t$, which guides $x_{t+1}$ via simulated degradation $y_t$. The right panel shows patch division, diffusion restoration, and stitching. (\textit{Bottom}) \TDiff performs 2$\times$ and 4$\times$ super-resolution, denoising, and deblurring, with inputs (left) and outputs (right) showing cleaner detail and preserved thermal boundaries.}
    \Description{Figure showing the \TDiff pipeline}

    \label{fig:intro-fig}
\end{figure}

Denoising Diffusion Probabilistic Models (DDPMs)~\cite{ho2020denoising} are strong generative priors for inverse imaging problems~\cite{kawar2022denoising, Zhu2023Denoising}, but their dependence on large datasets limits use in data-scarce domains like thermal imaging. To address these challenges, we propose a patch-based diffusion framework (\TDiff) that models local structure and serves as a learned prior for restoration. Training on small thermal patches captures localized degradations and increases data diversity, enabling effective learning under scarcity. At inference, full-resolution images are reconstructed by denoising overlapping patches and merging them with smooth spatial windowing to avoid artifacts and preserve continuity. To our knowledge, this is the first patch-based diffusion framework that serves as a learned plug-and-play prior for thermal image restoration across multiple tasks. Our main contributions are:

\begin{itemize}[leftmargin=*]
\item We propose a novel patch-based diffusion framework designed for thermal image degradations.  
\item We introduce a training strategy that addresses data scarcity through diverse patch-level sampling and robust merging.  
\item We demonstrate unified and competitive performance on denoising, super-resolution, and deblurring tasks.  
\end{itemize}

We validate \TDiff across key thermal restoration tasks, showing consistent gains over state-of-the-art methods and strong generalization to real consumer thermal cameras.

\section{Related Work}

\textbf{Solving Inverse Problems.} Image restoration is traditionally posed as an inverse problem, where the goal is to recover a clean image from a degraded observation. Traditional methods apply analytical priors like total variation~\cite{rudin1992nonlinear} or wavelet sparsity~\cite{donoho2002noising}, but often fall short under complex noise or blur. Learning-based methods, especially Convolutional Neural Networks, have demonstrated superior performance by directly learning mappings from degraded to clean images~\cite{dong2015image}. Hybrid frameworks like Plug-and-Play (PnP)~\cite{venkatakrishnan2013plug, romano2017little, zhang2017learning, zhang2021plug} incorporate learned denoisers within optimization algorithms, offering flexibility across tasks. However, these methods require large, high-quality datasets, often unavailable in thermal imaging.

\noindent\textbf{Diffusion-Based Restoration.} Denoising Diffusion Probabilistic Models (DDPMs)~\cite{ho2020denoising} have recently become dominant generative priors for image restoration~\cite{kawar2022denoising, Zhu2023Denoising}. Extensions like DiffIR~\cite{xia2023diffir} improve sampling efficiency and quality on RGB images using task-specific conditioning. WeatherDiffusion~\cite{weatherdiff} further introduces patch-based diffusion for RGB image restoration under adverse weather, showing the benefit of localized priors. However, these models are designed for visible light images and rely on large training corpora. In contrast, we propose a patch-based diffusion model designed for thermal restoration, where data is scarce and degradations are spatially local.

\noindent\textbf{Thermal Image Challenges and Restoration.} Thermal imaging is well suited for scenarios involving darkness, occlusion, or material classification~\cite{narayanan2024shape, sheinin2024projecting, wilson2023recent, dashpute2023thermal}, but presents distinct challenges such as low resolution, fixed pattern noise (FPN), and limited data. FPN appears as structured, spatially repetitive artifacts that models trained on RGB images often misinterpret as real features~\cite{zhang2017beyond, zhang2018ffdnet}. Public datasets remain scarce and temporally redundant, limiting generalization. Self-supervised~\cite{ulyanov2018deep, saragadam2021thermal} and physics-based~\cite{dashpute2023thermal} methods have been explored but generally underperform in high-noise or real-world conditions. Unlike prior work, we introduce a unified patch-based diffusion model that learns localized thermal priors and generalizes across denoising, super-resolution, and deblurring.

\section{Methodology}

We describe our approach for solving thermal inverse problems using patched-based denoising diffusion restoration. We begin with a brief overview of denoising diffusion models and their extension to inverse problems, and then present our patch-based framework adapted for thermal image restoration.

\subsection{Diffusion Models and Restoration}

Denoising Diffusion Probabilistic Models (DDPMs)~\cite{ho2020denoising} model the data distribution via a forward process that progressively adds Gaussian noise to a clean image $\mathbf{x}_0$ over $T$ timesteps. The reverse process, learned through a neural network $\epsilon_\theta(\mathbf{x}_t, t)$, predicts the noise at each timestep to iteratively recover the data. Sampling speed can be improved with DDIM~\cite{Song2020Denoising}, which introduces a non-Markovian formulation. To address inverse problems, DDRM~\cite{kawar2022denoising} modifies the reverse diffusion process to incorporate a linear degradation model:
\begin{equation}
\label{eq:linear_inverse_problem}
\mathbf{y} = \mathbf{A} \mathbf{x}_0 + \mathbf{n},
\end{equation}
where $\mathbf{A} \in \mathbb{R}^{m \times n}$ is the degradation operator and $\mathbf{n}$ is additive noise. To enforce alignment with the observation $\mathbf{y}$, each estimate $\hat{\mathbf{x}}_{0|t}$ is corrected using data-consistency guidance. Following algebraic reformulations of the projection step in~\cite{kawar2022denoising}, this correction can be expressed in two equivalent forms:

\begin{align}
\mathbf{g}_{\text{BP}}(\hat{\mathbf{x}}_{0|t}) &= \mathbf{A}^\top \left( \mathbf{A} \mathbf{A}^\top + \eta \mathbf{I} \right)^{-1} \left( \mathbf{A} \hat{\mathbf{x}}_{0|t} - \mathbf{y} \right), \label{eq:bp_guidance} \\
\mathbf{g}_{\text{LS}}(\hat{\mathbf{x}}_{0|t}) &= c \mathbf{A}^\top \left( \mathbf{A} \hat{\mathbf{x}}_{0|t} - \mathbf{y} \right), \label{eq:ls_guidance}
\end{align}

where $\eta$ is a regularization parameter and $c$ is a scalar weight. These guidance terms adjust the trajectory of the reverse process to stay consistent with the measured data.

\begin{figure*}[ht]
    \centering

    \begin{subfigure}{\textwidth}
        \centering
        \includegraphics[width=\textwidth]{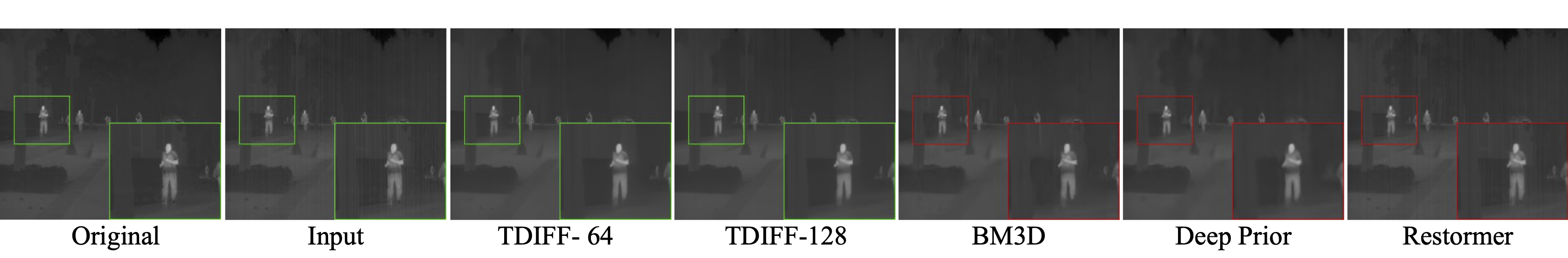}
        \caption{}
        \label{fig:denoise}
    \end{subfigure}

    \begin{subfigure}{\textwidth}
        \centering
        \includegraphics[width=\textwidth]{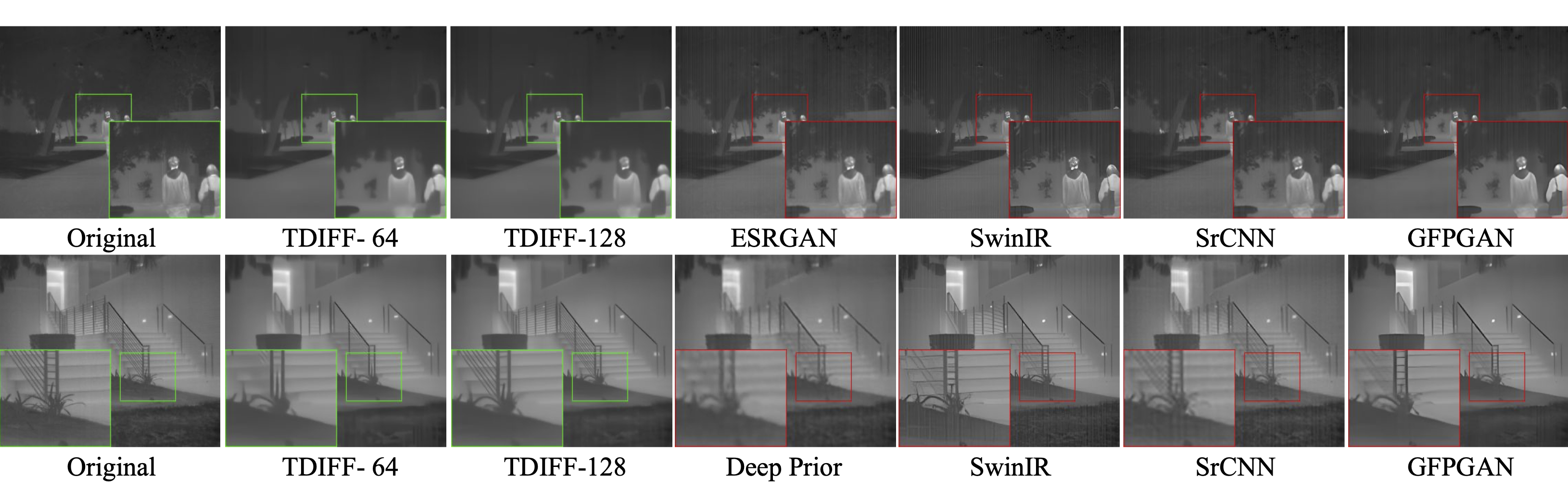}
        \caption{}
        \label{fig:superres}
    \end{subfigure}
    
    \begin{subfigure}{\textwidth}
        \centering
        \includegraphics[width=\textwidth]{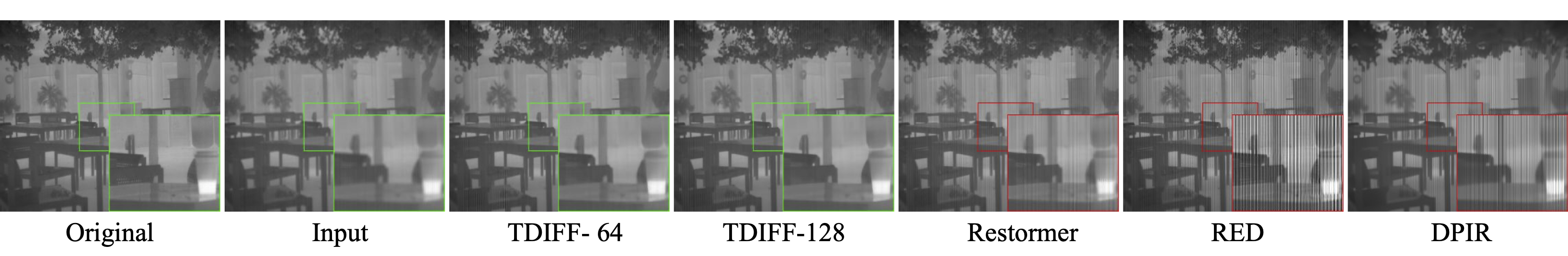}
        \caption{}
        \label{fig:deblur}
    \end{subfigure}

    \caption{Simulated FLIR results with added FPN. \TDiff outperforms baselines for (a) denoising, (b) 2$\times$ (top) and 4$\times$ (bottom) super-resolution, and (c) deblurring, preserving edges, suppressing noise, and maintaining thermal gradients.}

\label{fig:experiments}
    \Description{Figure simulated results}
\end{figure*}

\subsection{Patch-Based Diffusion for Thermal Images}
\label{sec:patch-diff}
Thermal datasets are generally limited in scale and diversity, which constrains the training of large generative models on full-resolution images. Moreover, degradations such as blur, downsampling, and fixed pattern noise often exhibit localized structure. To better capture these effects, we adopt a patch-based strategy inspired by~\cite{zhang2023unified}, training the diffusion model on $64 \times 64$ and $128 \times 128$ image crops. This increases the number of training examples, reduces memory requirements, and enables the model to learn statistical priors over edges, gradients, and material transitions~\cite{dong2015image}.

In our proposed approach, we divide the input image into overlapping patches, each of which is processed independently by the trained model. Directly merging the outputs can lead to seams and boundary artifacts due to inconsistencies across overlapping regions. To mitigate this, we perform patch division and recombination at every step of the reverse diffusion process. This iterative stitching strategy helps preserve spatial continuity throughout the denoising trajectory. Additionally, each patch is modulated by a smooth spatial window that suppresses edge artifacts and ensures consistency across overlaps during reconstruction.

Formally, at each timestep $t$, the current image estimate $\mathbf{x}_t$ is divided into overlapping patches of size $ps \times ps$ (with $ps \in {64, 128}$). Each patch is denoised independently to produce intermediate predictions $\hat{\mathbf{x}}_{0|t,k}$. These predictions are aggregated into a full image using spatially weighted averaging:
\begin{equation}
\hat{\mathbf{x}}(i,j) = \sum_k w_k(i,j) \hat{\mathbf{x}}_{0|t,k}(i,j),
\end{equation}
where $w_k(i,j)$ denotes the window function applied to the $k$th patch. The resulting image $\hat{\mathbf{x}}_t$ is then refined by incorporating measurement guidance through a weighted combination of back-projection (BP) and least squares (LS) correction:
\begin{equation}
\mathbf{x}_{t-1} = \hat{\mathbf{x}}_t - \mu_t \left( (1 - \delta_t) \mathbf{g}_{\text{BP}} + \delta_t \mathbf{g}_{\text{LS}} \right),
\end{equation}
where $\mu_t$ is the step size and $\delta_t \in [0,1]$ controls the balance between the two guidance terms.

\section{Implementation Details}

We trained our models on the FLIR Thermal Dataset~\cite{FLIR2019ADK}, which consists of 10{,}742 grayscale thermal images at a resolution of $640 \times 512$. Training samples were generated by randomly cropping patches of size $32 \times 32$, $64 \times 64$, and $128 \times 128$. Section~\ref{sec:ablation} presents a qualitative comparison of patch size effects. All patches were sampled directly from the original images to preserve the native structure and statistics of thermal noise, including fixed pattern artifacts. Low-variance patches were excluded to ensure meaningful content in each sample. The denoising model uses a U-Net architecture~\cite{ronneberger2015u} and is trained with the standard DDPM objective, minimizing mean squared error between added noise and the model's prediction at each diffusion step. Our implementation is based on the official DDPM~\cite{ho2020denoising} and DDRM~\cite{kawar2022denoising} repositories. We follow the DDPM training framework and use DDRM for inference. Both pipelines were adapted for grayscale thermal images and modified to support patch-based sampling, overlapping reconstruction, and measurement-guided reverse diffusion as detailed in Section~\ref{sec:patch-diff}. Additional implementation details are provided in Section~\ref{sec:supp-implementation} of the Supplementary.

\section{Experiments}

We evaluate our patch-based thermal diffusion method (\TDiff) on three inverse problems, including denoising, super-resolution, and deblurring, using simulated and real-world thermal data, with real-world results presented in Section~\ref{sec:supp-real-world} of the Supplementary. Our approach is compared with both classical and state-of-the-art restoration methods.

\begin{table}[ht]
  \centering
  \caption{Denoising results on the FLIR validation set. Best results are in \textbf{bold}. Our method achieves the highest SSIM, reflecting superior structural preservation.}
  \label{tab:denoising}
  \begin{tabular}{@{}lcc@{}}
    \toprule
    Method & PSNR (dB) & SSIM \\
    \midrule
    \TDiffST (ours) & 35.8 & \textbf{0.959} \\
    \TDiffLT (ours) & 35.7 & 0.954 \\
    BM3D~\cite{dabov2007image} & 36.2 & 0.956 \\
    Restormer~\cite{zamir2022restormer} & 35.9 & 0.952 \\
    NAFNet~\cite{chen2022simple} & 35.5 & 0.952 \\
    MIRNet~\cite{zamir2020learning} & 35.2 & 0.950 \\
    FFDNet~\cite{zhang2018ffdnet} & 35.0 & 0.949 \\
    FDnCNN~\cite{zhang2017beyond} & 34.9 & 0.943 \\
    Deep Image Prior~\cite{ulyanov2018deep} & \textbf{37.1} & 0.945 \\
    GFPGAN~\cite{wang2021towards} & 34.5 & 0.897 \\
    DiffIR~\cite{xia2023diffir} & 30.4 & 0.806 \\
    WeatherDiffusion~\cite{weatherdiff} & 20.6 & 0.805 \\
    \bottomrule
  \end{tabular}
\end{table}

\subsection{Simulated Inverse Problems}

All simulations were performed on the FLIR validation dataset with added fixed pattern noise (FPN) to mimic realistic thermal degradation.

\subsubsection{Denoising}
Results in Table~\ref{tab:denoising} show that \TDiffS and \TDiffL achieve SSIM scores of 0.959 and 0.954, respectively, with \TDiffS outperforming strong baselines such as BM3D and Restormer. Visual comparisons in Figure~\ref{fig:experiments}(a) further demonstrate that our models more effectively preserve structural edges, suppress noise, and maintain thermal gradients. While Deep Image Prior reports the highest PSNR, this is largely due to its tendency to overfit to the noisy input, which can lead to high pixel-wise similarity but degrades perceptual quality by introducing artifacts and suppressing fine structures. GAN-based methods such as GFPGAN fail to generalize to thermal noise and often produce unnatural textures inconsistent with the thermal image statistics.

\begin{table}[!tt]
  \centering
  \caption{2$\times$ super-resolution results on the FLIR validation set. Best results are in \textbf{bold}. Our method achieves the highest SSIM, indicating superior structural preservation.}
  \label{tab:sr2}
  \begin{tabular}{@{}lcc@{}}
    \toprule
    Method & PSNR (dB) & SSIM \\
    \midrule
    \TDiffST (ours) & 35.3 & \textbf{0.936} \\
    \TDiffLT (ours) & 35.2 & 0.934 \\
    SRCNN~\cite{dong2015image} & \textbf{36.1} & 0.928 \\
    GFPGAN~\cite{wang2021towards} & 34.9 & 0.901 \\
    SwinIR~\cite{liang2021swinir} & 32.0 & 0.817 \\
    ESRGAN~\cite{wang2018esrgan} & 28.3 & 0.736 \\
    \bottomrule
  \end{tabular}
\end{table}

\begin{figure}[ht]
\centering
\includegraphics[width=\columnwidth]{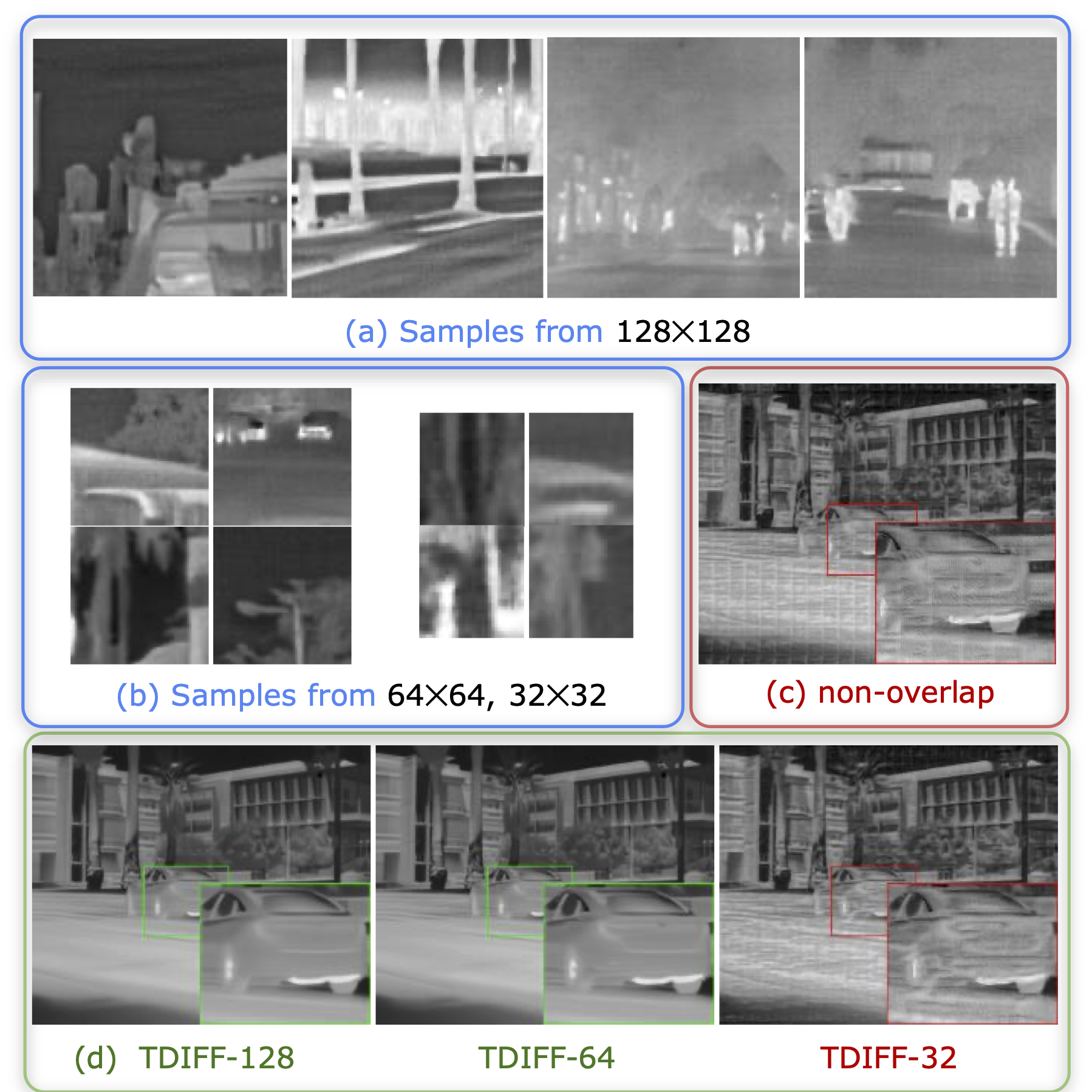}
\caption{Impact of patch size and overlap on restoration quality. (a, b) Smaller patch sizes reduce unconditional sample quality. (c) Non-overlapping reconstruction leads to visible seams. (d) \TDiff variants show trade-offs between patch size and restoration quality.}
\Description{path size effect}
\label{fig:ddpm-sample} 
\end{figure}

\begin{table}[!tt]
  \centering
  \caption{4$\times$ super-resolution results on the FLIR validation set. Best results are in \textbf{bold}. Our method achieves the highest SSIM, reflecting superior structural preservation.}
  \label{tab:sr4}
  \begin{tabular}{@{}lcc@{}}
    \toprule
    Method & PSNR (dB) & SSIM \\
    \midrule
    \TDiffST (ours) & 34.1 & 0.876 \\
    \TDiffLT(ours) & \textbf{34.9} & \textbf{0.887} \\
    SRCNN~\cite{dong2015image} & 34.4 & 0.881 \\
    GFPGAN~\cite{wang2021towards} & 32.7 & 0.869 \\
    Deep Image Prior~\cite{ulyanov2018deep} & 33.2 & 0.843 \\
    SwinIR~\cite{liang2021swinir} & 31.3 & 0.792 \\
    \bottomrule
  \end{tabular}
\end{table}

\begin{table}[!tt]
  \centering
  \caption{Deblurring results on the FLIR validation set. Best results are in \textbf{bold}. Our method achieves the highest SSIM and PSNR, showing superior perceptual and pixel-level fidelity.}
  \label{tab:deblur}
  \begin{tabular}{@{}lcc@{}}
    \toprule
    Method & PSNR (dB) & SSIM \\
    \midrule
    \TDiffST (ours) & 32.5 & 0.864 \\
    \TDiffLT (ours)  & \textbf{33.6} & \textbf{0.866} \\
    Restormer~\cite{zamir2022restormer} & 30.5 & 0.792 \\
    DPIR~\cite{zhang2021plug} & 22.0 & 0.500 \\
    RED~\cite{mao2016image} & 20.2 & 0.489 \\
    \bottomrule
  \end{tabular}
\end{table}

\subsubsection{Super-Resolution}
As seen in Table~\ref{tab:sr2}, \TDiffS and \TDiffL achieve SSIM scores of 0.936 and 0.934 at 2$\times$ upscaling, outperforming all baselines in structural fidelity. While SRCNN yields the highest PSNR, it oversmooths, sacrificing perceptual detail. In contrast, our models preserve texture and structure, as reflected in SSIM and Figure~\ref{fig:experiments}(b). Table~\ref{tab:sr4} shows the results for 4$\times$ upscaling, where \TDiffL achieves the highest PSNR of 34.9 and SSIM of 0.887, benefiting from its larger receptive field. Figure~\ref{fig:experiments}(b) shows that GANs introduce artifacts and SwinIR fails to preserve thermal continuity.

\subsubsection{Deblurring}
Table~\ref{tab:deblur} shows that \TDiffL achieves the highest performance, with 33.6\,dB PSNR and 0.866 SSIM, surpassing both classical and learning-based baselines. As shown in Figure~\ref{fig:experiments}(c), \TDiff restores sharper edges and clearer structural details compared to DPIR, RED, and Restormer, which either fail to remove blur or amplify fixed pattern noise.

\subsection{Ablation Studies}
\label{sec:ablation}
We investigated the impact of patch size and overlap on restoration quality, as shown in Figure~\ref{fig:ddpm-sample}. Figures~\ref{fig:ddpm-sample}(a, b) show that models trained with small patches (e.g., $32 \times 32$) failed to capture sufficient context, leading to lower quality samples and poor global structure reconstruction. In contrast, $64 \times 64$ and $128 \times 128$ patches produced more coherent and detailed outputs. Figure~\ref{fig:ddpm-sample}(c) illustrates how stitching non-overlapping patches creates seams and grid-like artifacts, particularly in regions with smooth thermal transitions. Overlapping patches combined with spatial windowing help reduce these artifacts, leading to cleaner, more consistent outputs. As shown in Figure~\ref{fig:ddpm-sample}(d), models trained on larger patches produce noticeably better restorations, reinforcing our decision to use $64 \times 64$ and $128 \times 128$ patches in all experiments. Additional runtime comparisons are provided in Section~\ref{sec:supp-implementation} of the Supplementary.

\section{Conclusion}
We introduced a patch-based diffusion framework for thermal image restoration that leverages the local nature of degradations like noise, blur, and resolution loss. We demonstrated that training with $64 \times 64$ and $128 \times 128$ image patches enables the model to learn fine structural priors while maintaining efficiency in data-limited settings. In this approach, full images are restored by processing overlapping patches and merging outputs with smooth blending. Our method outperforms classical and learning-based baselines across denoising, super-resolution, and deblurring on both simulated and real thermal data. Future work includes blind restoration and extensions to MWIR, hyperspectral, and event-based imaging. Our code will be publicly released for reproducibility.

\balance

\bibliographystyle{unsrt}
\bibliography{ref}

\begin{thebibliography}{10}

\bibitem{wilson2023recent}
AN~Wilson, Khushi~Anil Gupta, Balu~Harshavardan Koduru, Abhinav Kumar, Ajit Jha, and Linga~Reddy Cenkeramaddi.
\newblock Recent advances in thermal imaging and its applications using machine learning: A review.
\newblock {\em IEEE Sensors Journal}, 23(4):3395--3407, 2023.

\bibitem{dashpute2023thermal}
Aniket Dashpute, Vishwanath Saragadam, Emma Alexander, Florian Willomitzer, Aggelos Katsaggelos, Ashok Veeraraghavan, and Oliver Cossairt.
\newblock Thermal spread functions (tsf): Physics-guided material classification.
\newblock In {\em Proceedings of the IEEE/CVF Conference on Computer Vision and Pattern Recognition}, pages 1641--1650, 2023.

\bibitem{venkatakrishnan2013plug}
Singanallur~V Venkatakrishnan, Charles~A Bouman, and Brendt Wohlberg.
\newblock Plug-and-play priors for model based reconstruction.
\newblock In {\em 2013 IEEE global conference on signal and information processing}, pages 945--948. IEEE, 2013.

\bibitem{romano2017little}
Yaniv Romano, Michael Elad, and Peyman Milanfar.
\newblock The little engine that could: Regularization by denoising (red).
\newblock {\em SIAM journal on imaging sciences}, 10(4):1804--1844, 2017.

\bibitem{zhang2017beyond}
Kai Zhang, Wangmeng Zuo, Yunjin Chen, Deyu Meng, and Lei Zhang.
\newblock Beyond a gaussian denoiser: Residual learning of deep cnn for image denoising.
\newblock {\em IEEE transactions on image processing}, 26(7):3142--3155, 2017.

\bibitem{zhang2018ffdnet}
Kai Zhang, Wangmeng Zuo, and Lei Zhang.
\newblock Ffdnet: Toward a fast and flexible solution for cnn-based image denoising.
\newblock {\em IEEE Transactions on Image Processing}, 27(9):4608--4622, 2018.

\bibitem{rudin1992nonlinear}
Leonid~I Rudin, Stanley Osher, and Emad Fatemi.
\newblock Nonlinear total variation based noise removal algorithms.
\newblock {\em Physica D: nonlinear phenomena}, 60(1-4):259--268, 1992.

\bibitem{donoho2002noising}
David~L Donoho.
\newblock De-noising by soft-thresholding.
\newblock {\em IEEE transactions on information theory}, 41(3):613--627, 2002.

\bibitem{ulyanov2018deep}
Dmitry Ulyanov, Andrea Vedaldi, and Victor Lempitsky.
\newblock Deep image prior.
\newblock In {\em Proceedings of the IEEE conference on computer vision and pattern recognition}, pages 9446--9454, 2018.

\bibitem{saragadam2021thermal}
Vishwanath Saragadam, Akshat Dave, Ashok Veeraraghavan, and Richard~G Baraniuk.
\newblock Thermal image processing via physics-inspired deep networks.
\newblock In {\em Proceedings of the IEEE/CVF International Conference on Computer Vision}, pages 4057--4065, 2021.

\bibitem{ho2020denoising}
Jonathan Ho, Ajay Jain, and Pieter Abbeel.
\newblock Denoising diffusion probabilistic models.
\newblock {\em Advances in neural information processing systems}, 33:6840--6851, 2020.

\bibitem{kawar2022denoising}
Bahjat Kawar, Michael Elad, Stefano Ermon, and Jiaming Song.
\newblock Denoising diffusion restoration models.
\newblock {\em Advances in neural information processing systems}, 35:23593--23606, 2022.

\bibitem{Zhu2023Denoising}
Yuanzhi Zhu, Kai Zhang, Jingyun Liang, Jiezhang Cao, Bihan Wen, Radu Timofte, and Luc Van~Gool.
\newblock Denoising diffusion models for plug-and-play image restoration.
\newblock In {\em Proceedings of the IEEE/CVF conference on computer vision and pattern recognition}, pages 1219--1229, 2023.

\bibitem{dong2015image}
Chao Dong, Chen~Change Loy, Kaiming He, and Xiaoou Tang.
\newblock Image super-resolution using deep convolutional networks.
\newblock {\em IEEE transactions on pattern analysis and machine intelligence}, 38(2):295--307, 2015.

\bibitem{zhang2017learning}
Kai Zhang, Wangmeng Zuo, Shuhang Gu, and Lei Zhang.
\newblock Learning deep cnn denoiser prior for image restoration.
\newblock In {\em Proceedings of the IEEE conference on computer vision and pattern recognition}, pages 3929--3938, 2017.

\bibitem{zhang2021plug}
Kai Zhang, Yawei Li, Wangmeng Zuo, Lei Zhang, Luc Van~Gool, and Radu Timofte.
\newblock Plug-and-play image restoration with deep denoiser prior.
\newblock {\em IEEE Transactions on Pattern Analysis and Machine Intelligence}, 44(10):6360--6376, 2021.

\bibitem{xia2023diffir}
Bin Xia, Yulun Zhang, Shiyin Wang, Yitong Wang, Xinglong Wu, Yapeng Tian, Wenming Yang, and Luc Van~Gool.
\newblock Diffir: Efficient diffusion model for image restoration.
\newblock In {\em Proceedings of the IEEE/CVF international conference on computer vision}, pages 13095--13105, 2023.

\bibitem{weatherdiff}
Ozan Özdenizci and Robert Legenstein.
\newblock Restoring vision in adverse weather conditions with patch-based denoising diffusion models.
\newblock {\em IEEE Transactions on Pattern Analysis and Machine Intelligence}, 45(8):10346--10357, 2023.

\bibitem{narayanan2024shape}
Sriram Narayanan, Mani Ramanagopal, Mark Sheinin, Aswin~C Sankaranarayanan, and Srinivasa~G Narasimhan.
\newblock Shape from heat conduction.
\newblock In {\em European Conference on Computer Vision}, pages 426--444. Springer, 2024.

\bibitem{sheinin2024projecting}
Mark Sheinin, Aswin~C Sankaranarayanan, and Srinivasa~G Narasimhan.
\newblock Projecting trackable thermal patterns for dynamic computer vision.
\newblock In {\em Proceedings of the IEEE/CVF Conference on Computer Vision and Pattern Recognition}, pages 25223--25232, 2024.

\bibitem{Song2020Denoising}
Jiaming Song, Chenlin Meng, and Stefano Ermon.
\newblock Denoising diffusion implicit models.
\newblock {\em arXiv preprint arXiv:2010.02502}, 2020.

\bibitem{zhang2023unified}
Yi~Zhang, Xiaoyu Shi, Dasong Li, Xiaogang Wang, Jian Wang, and Hongsheng Li.
\newblock A unified conditional framework for diffusion-based image restoration.
\newblock {\em Advances in Neural Information Processing Systems}, 36:49703--49714, 2023.

\bibitem{FLIR2019ADK}
{FLIR Systems Inc.}
\newblock Flir thermal dataset for algorithm training.
\newblock \url{https://www.flir.com/oem/adas/adas-dataset-form/}, 2019.

\bibitem{ronneberger2015u}
Olaf Ronneberger, Philipp Fischer, and Thomas Brox.
\newblock U-net: Convolutional networks for biomedical image segmentation.
\newblock In {\em International Conference on Medical image computing and computer-assisted intervention}, pages 234--241. Springer, 2015.

\bibitem{dabov2007image}
Kostadin Dabov, Alessandro Foi, Vladimir Katkovnik, and Karen Egiazarian.
\newblock Image denoising by sparse 3-d transform-domain collaborative filtering.
\newblock {\em IEEE Transactions on image processing}, 16(8):2080--2095, 2007.

\bibitem{zamir2022restormer}
Syed~Waqas Zamir, Aditya Arora, Salman Khan, Munawar Hayat, Fahad~Shahbaz Khan, and Ming-Hsuan Yang.
\newblock Restormer: Efficient transformer for high-resolution image restoration.
\newblock In {\em Proceedings of the IEEE/CVF conference on computer vision and pattern recognition}, pages 5728--5739, 2022.

\bibitem{chen2022simple}
Liangyu Chen, Xiaojie Chu, Xiangyu Zhang, and Jian Sun.
\newblock Simple baselines for image restoration.
\newblock In {\em European conference on computer vision}, pages 17--33. Springer, 2022.

\bibitem{zamir2020learning}
Syed~Waqas Zamir, Aditya Arora, Salman Khan, Munawar Hayat, Fahad~Shahbaz Khan, Ming-Hsuan Yang, and Ling Shao.
\newblock Learning enriched features for real image restoration and enhancement.
\newblock In {\em European conference on computer vision}, pages 492--511. Springer, 2020.

\bibitem{wang2021towards}
Xintao Wang, Yu~Li, Honglun Zhang, and Ying Shan.
\newblock Towards real-world blind face restoration with generative facial prior.
\newblock In {\em Proceedings of the IEEE/CVF conference on computer vision and pattern recognition}, pages 9168--9178, 2021.

\bibitem{liang2021swinir}
Jingyun Liang, Jiezhang Cao, Guolei Sun, Kai Zhang, Luc Van~Gool, and Radu Timofte.
\newblock Swinir: Image restoration using swin transformer.
\newblock In {\em Proceedings of the IEEE/CVF international conference on computer vision}, pages 1833--1844, 2021.

\bibitem{wang2018esrgan}
Xintao Wang, Ke~Yu, Shixiang Wu, Jinjin Gu, Yihao Liu, Chao Dong, Yu~Qiao, and Chen Change~Loy.
\newblock Esrgan: Enhanced super-resolution generative adversarial networks.
\newblock In {\em Proceedings of the European conference on computer vision (ECCV) workshops}, pages 0--0, 2018.

\bibitem{mao2016image}
Xiaojiao Mao, Chunhua Shen, and Yu-Bin Yang.
\newblock Image restoration using very deep convolutional encoder-decoder networks with symmetric skip connections.
\newblock {\em Advances in neural information processing systems}, 29, 2016.

\end{thebibliography}

\clearpage    
\appendix
\section*{Supplementary}
\label{sec:supplementary}

\section{Real-World Experiments}
\label{sec:supp-real-world}

We validated \TDiff on real thermal data captured by two cameras. The FLIR Boson+ camera ($640 \times 512$, NETD 20,mK) was used for denoising, and the Seek Mosaic ($320 \times 240$, NETD 60,mK) for super-resolution.

\begin{figure*}[ht]
    \centering
    \begin{subfigure}{0.9\textwidth}
        \centering
        \includegraphics[width=0.9\textwidth]{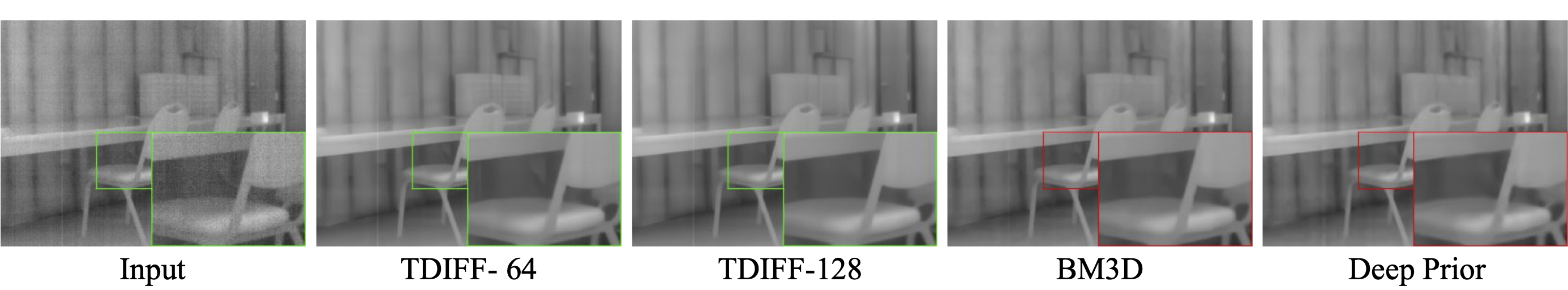}
        \caption{}
        \label{fig:denoise-real}
    \end{subfigure}
        
    \begin{subfigure}{0.9\textwidth}
        \centering
        \includegraphics[width=0.9\textwidth]{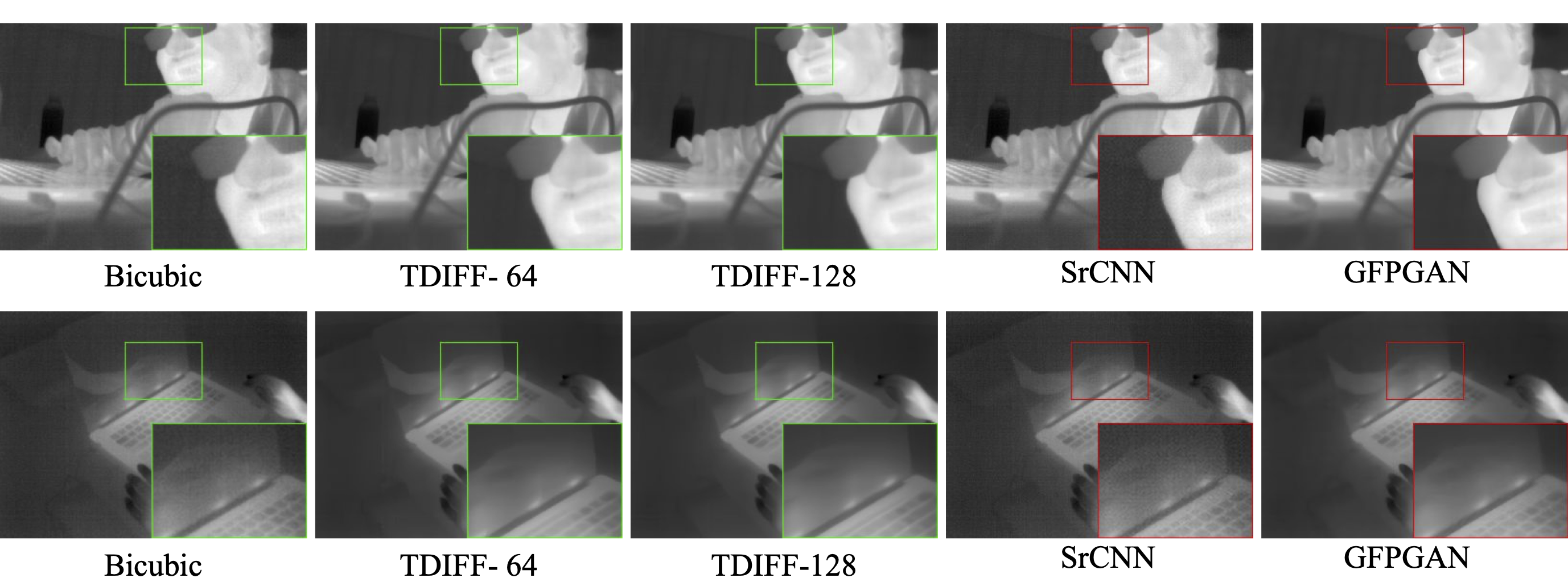}
        \caption{}
        \label{fig:superres-real}
    \end{subfigure}
    
    \caption{Real-world thermal image restoration results, comparing our method to competing approaches. (a) Shows denoising results on FLIR Boson+ data, (b) presents 2$\times$ (top) and 4$\times$ (bottom) super-resolution results on Seek Mosaic data. Our models produce cleaner outputs with preserved structure and thermal continuity, effectively suppressing noise and artifacts across both tasks.}   
    \label{fig:experiments-real}
    \Description{Figure real world results}
\end{figure*}

\subsection{Denoising}
Figure~\ref{fig:experiments-real}(a) illustrates the denoising results on real thermal data. Classical methods such as BM3D~\cite{dabov2007image} suppress noise but blur structural edges and weaken thermal gradients. Compared to Deep Image Prior~\cite{ulyanov2018deep}, our models produce visually cleaner outputs with better preservation of structure and thermal continuity. \TDiffL and \TDiffS maintain fine structural details while effectively suppressing both fixed pattern and random thermal noise. Table~\ref{tab:supp-denoising} presents the denoising results, where \TDiffL achieved the highest SSIM (0.922) and \TDiffS closely matched the best PSNR while maintaining high structural fidelity. Although BM3D~\cite{dabov2007image} attained slightly higher PSNR, it blurred structural edges and weakened thermal gradients, resulting in lower SSIM, as illustrated in Figure~\ref{fig:experiments-real}(a).

\subsection{Super-Resolution}
Figure~\ref{fig:experiments-real}(b) shows super-resolution results on real-world thermal images. \TDiff models consistently enhanced resolution while preserving natural thermal textures and suppressing upsampling artifacts. GAN-based methods such as GFPGAN~\cite{wang2021towards} introduced unrealistic textures, and models like SrCNN~\cite{dong2015image} failed to recover thermal continuity. \TDiff provided sharper, artifact-free reconstructions with consistent thermal structure.

\section{Patch-Based Training Setup}
\label{sec:supp-implementation}
We used the PyTorch implementation of Denoising Diffusion Probabilistic Model (DDPM) ~\cite{ho2020denoising} for single-channel grayscale thermal image restoration. The denoising network is a U-Net architecture~\cite{ronneberger2015u} with configurations determined by patch size.

\begin{table}[!b]
  \centering
  \caption{Denoising results on real thermal data from the FLIR Boson+ camera. The best result for each metric is shown in \textbf{bold}. Our approach achieves the highest SSIM, indicating superior structural preservation.}
  \label{tab:supp-denoising}
  \begin{tabular}{@{}lcc@{}}
    \toprule
    Method & PSNR (dB) & SSIM \\
    \midrule
    \TDiffST (ours) & 36.46 & 0.916 \\
    \TDiffLT (ours) & 36.55 & \textbf{0.922} \\
    Restormer~\cite{zamir2022restormer} & 35.9 & 0.908 \\
    BM3D~\cite{dabov2007image} & \textbf{36.6} & 0.908 \\
    FDnCNN~\cite{zhang2017beyond} & 35.5 & 0.900 \\
    \bottomrule
  \end{tabular}
\end{table}

For $128 \times 128$ patches, the base channel count is 64 with channel multipliers [1, 1, 2, 2, 4, 4]. For $64 \times 64$ patches, the base channel count is reduced to 32 while keeping the same multipliers, providing a balanced architecture that preserves capacity while reducing computation. The noise schedule is linear with $\beta_{\text{start}} = 0.0001$ and $\beta_{\text{end}} = 0.02$ over 1000 diffusion steps, producing a gradual denoising process from heavily corrupted to clean images. Training uses the Adam optimizer with a learning rate of 0.0002, which adjusts parameter-specific learning rates to improve convergence stability.

For patch selection, we randomly sample $128 \times 128$ and $64 \times 64$ crops from the FLIR dataset ~\cite{FLIR2019ADK}, compute their variance, and retain only those above 0.6 to ensure informative content and diversity. Training batches contain 192 patches for the $128 \times 128$ configuration and 384 for the $64 \times 64$ configuration, and models are trained for 10{,}000 epochs on NVIDIA RTX A5000 GPUs. This patch-based strategy increases the effective training sample size, captures diverse thermal features, and enables learning of robust priors despite limited dataset availability.

\subsection{Hyperparameter Tuning}

\begin{figure}[H]
    \centering
    \includegraphics[width=\columnwidth]{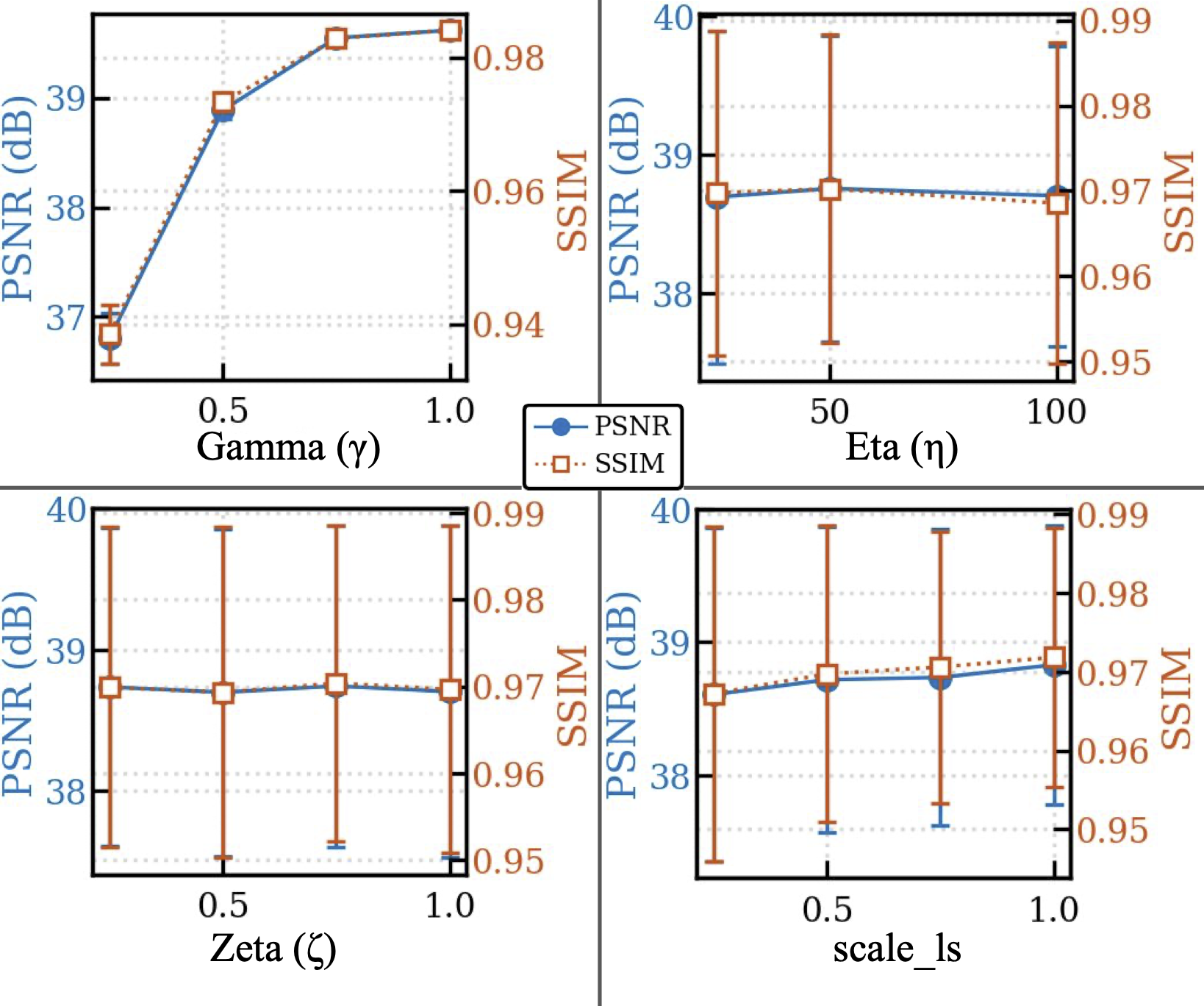}
    \caption{Quantitative comparison of hyperparameter settings for denoising performance.}
    \label{fig:hyper_param}
    \Description{Figure hyperparams}
\end{figure}

We conducted ablation studies to assess the impact of key hyperparameters on the performance of our \TDiff models. Specifically, we examined the effects of the scaling factor $\gamma$, the regularization parameter $\eta$, and the blending weight $\zeta$ on restoration quality. The results are summarized in Figure~\ref{fig:hyper_param}, which presents a quantitative comparison of how each hyperparameter influences denoising performance.

\subsubsection{Scaling Factor}
$\text{scale}{\text{LS}}$ is a multiplicative factor applied to the least-squares (LS) guidance term in the update equation, controlling the influence of data consistency. Increasing $\text{scale}{\text{LS}}$ strengthens the fit to the observed data, which can preserve fine details but may also amplify noise if it is present. Decreasing $\text{scale}{\text{LS}}$ reduces this influence, often producing smoother images that deviate from the observations. We observed a PSNR improvement from $37.0$,dB to $39.5$,dB as $\text{scale}{\text{LS}}$ increased from $0.2$ to $1.0$. The optimal range was between $0.8$ and $1.0$, where performance gains plateaued, balancing computational efficiency and image quality.

\subsubsection{Gamma ($\gamma$)}
$\gamma$ controls the balance between the data fidelity term and the prior term in the loss function during the diffusion process. Increasing $\gamma$ emphasizes the learned prior, producing smoother images but with potential oversmoothing and loss of important details. Decreasing $\gamma$ emphasizes data fidelity, preserving details but potentially retaining more noise or artifacts. We found that setting $\gamma$ between $60$ and $100$ provided an effective trade-off, reducing noise while preserving critical thermal details without overfitting to the noise in the observations.

\subsubsection{Eta ($\eta$)}
$\eta$ controls the amount of noise injected during the reverse diffusion process, modulating the stochasticity of image reconstruction. 
Higher $\eta$ values introduce more randomness, which can enhance diversity in generated images but may also increase blurriness and instability. Lower $\eta$ values produce sharper and more detailed images but reduce diversity and can increase the risk of overfitting. In our experiments, $\eta$ values between $0.6$ and $0.8$ achieved the best trade-off, producing sharp, detailed outputs without excessive artifacts or instability.

\subsubsection{Zeta ($\zeta$)}
$\zeta$ determines the trade-off between the model’s prediction and the observed data during the reverse diffusion process. Increasing $\zeta$ shifts emphasis toward the model’s estimation, producing outputs that align more closely with learned patterns but may diverge from the observed data. Decreasing $\zeta$ places more weight on the observed data, preserving its details but potentially retaining noise. Values between $0.8$ and $1.0$ yielded the highest reconstruction quality while maintaining consistency with the observed thermal data, balancing smoothness and data fidelity.

\subsection{Impact of Patch Size}
\begin{figure}[ht]
    \centering
    \includegraphics[width=\columnwidth]{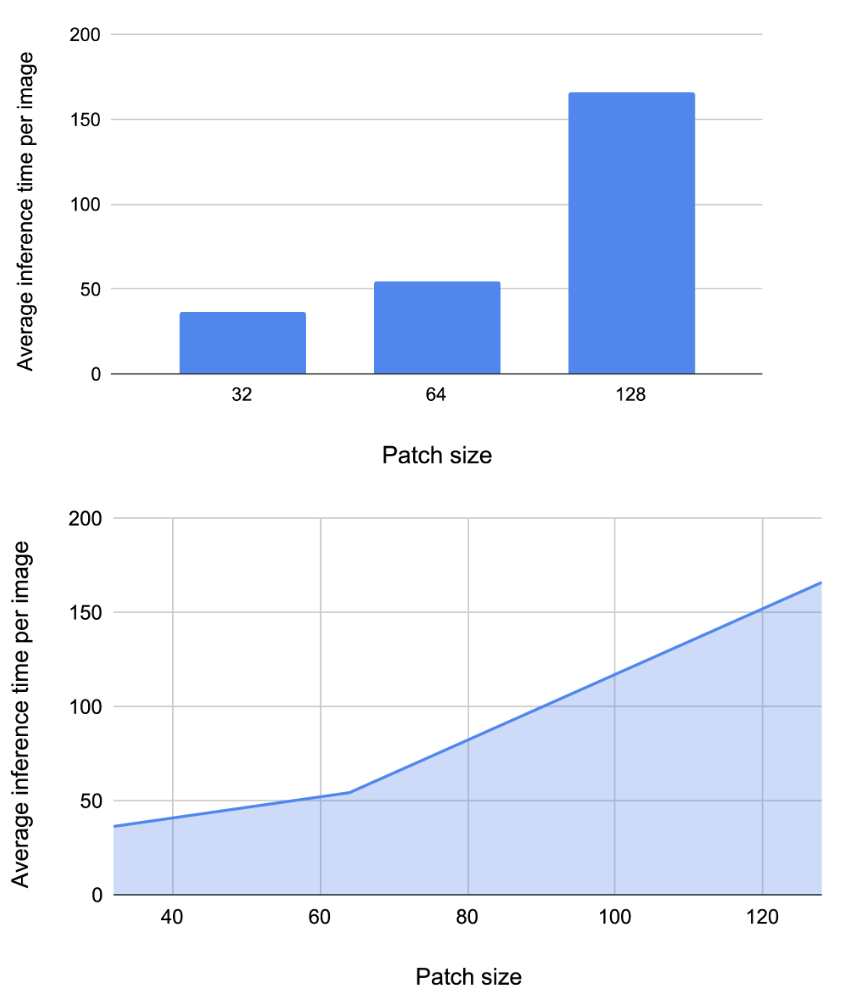}
    \caption{Average inference time of \TDiff in seconds for different patch sizes. Up: bar chart showing average time per image. Down: line plot illustrating the scaling trend. Larger patches improve quality but significantly increase computational cost.}
    \label{fig:patch-size-time}
    \Description{Figure timing}
\end{figure}

We conducted an ablation study to examine the relationship between patch size and inference time in \TDiff, as shown in Figure~\ref{fig:patch-size-time}. Average times per image were 36.23\,s for $32 \times 32$ patches, 54.21\,s for $64 \times 64$ patches, and 165.89\,s for $128 \times 128$ patches. As expected, larger patches improve restoration quality by providing more context but also incur substantially higher computational costs.

\subsection{Impact of Patch Overlap}

The importance of overlapping patches is illustrated in Figure~\ref{fig:ddpm-sample}(c). Removing patch overlap introduces visible discontinuities along patch boundaries, appearing as sharp, grid-like seams in the reconstructed image. These artifacts are most apparent in regions with gradual thermal transitions, where abrupt changes between adjacent patches disrupt smooth gradients. This observation supports our use of overlapping patches with appropriate blending, which ensures seamless reconstruction across the thermal image and avoids artificial boundaries. The overlapping strategy is particularly critical for thermal imaging, where maintaining continuous temperature gradients is essential for accurate interpretation.

\end{document}